\documentclass[conference]{IEEEtran}
\IEEEoverridecommandlockouts
\usepackage{cite}
\usepackage{amsmath,amssymb,amsfonts}
\usepackage{algorithmic}
\usepackage{graphicx}
\usepackage{textcomp}
\usepackage{xcolor}
\usepackage{hyperref}
\usepackage{url}
\usepackage[utf8]{inputenc}   % tells LaTeX your file is saved in UTF-8
\usepackage[T1]{fontenc}      % proper output encoding for accented characters
\def\BibTeX{{\rm B\kern-.05em{\sc i\kern-.025em b}\kern-.08em
    T\kern-.1667em\lower.7ex\hbox{E}\kern-.125emX}}
\begin{document}

\title{Evaluating Fine-Tuned LLM Model For Medical Transcription With Small Low-Resource Languages Validated Dataset \\
% {\footnotesize \textsuperscript{*}Note: Sub-titles are not captured for https://ieeexplore.ieee.org  and
% should not be used}
\thanks{M.N-R. Chowdhury is with the School of ICT and Industrial Management at Metropolia University of Applied Sciences, Helsinki, Finland (e-mail: nowshad.cse@gmail.com). This work was completed by Mohammed Nowshad Ruhani Chowdhury as part of the Metropolia University of Applied Sciences Research project: Digital scribes. No external funding was received. }
}

\author{
\IEEEauthorblockN{Mohammed Nowshad Ruhani\\ Chowdhury}
\IEEEauthorblockA{School of ICT and Industrial Management}
\IEEEauthorblockA{
\textit{Metropolia University of Applied Sciences}\\
Helsinki, Finland \\
mohammed.chowdhury@metropolia.fi \\
0000-0002-7429-1483}

\and
\IEEEauthorblockN{Mohammed Nowaz Rabbabi\\ Chowdhury}
\IEEEauthorblockA{Department of Electrical, Computer, \\and Systems Engineering}
\IEEEauthorblockA{
\textit{Rensselaer Polytechnic Institute}\\
Troy, NY, USA\\
chowdm2@rpi.edu}
\and

\centering
\IEEEauthorblockN{Sakari Lukkarinen}
\IEEEauthorblockA{School of ICT and Industrial Management}
\IEEEauthorblockA{
\textit{Metropolia University of Applied Sciences}\\
Helsinki, Finland \\
sakari.lukkarinen@metropolia.fi \\
0009-0007-2789-0049}
}

\maketitle

\begin{abstract}

Clinical documentation is a critical factor for patient safety, diagnosis, and continuity of care. The administrative burden of EHRs is a significant factor in physician burnout. This is a critical issue for low-resource languages, including Finnish. This study aims to investigate the effectiveness of a domain-aligned natural language processing (NLP); large language model for medical transcription in Finnish by fine-tuning LLaMA 3.1-8B on a small validated corpus of simulated clinical conversations by students at Metropolia University of Applied Sciences. The fine-tuning process for medical transcription used a controlled preprocessing and optimization approach. The fine-tuning effectiveness was evaluated by sevenfold cross-validation. The evaluation metrics for fine-tuned LLaMA 3.1-8B were BLEU = 0.1214, ROUGE-L = 0.4982, and BERTScore F1 = 0.8230. The results showed a low n-gram overlap but a strong semantic similarity with reference transcripts. This study indicate that fine-tuning can be an effective approach for translation of medical discourse in spoken Finnish and support the feasibility of fine-tuning a privacy-oriented domain-specific large language model for clinical documentation in Finnish. Beside that provide directions for future work.

% In modern healthcare, clinical documentation is paramount for patient safety, accurate diagnoses, and continuity of care. However, physician burnout has been caused by the increasing overhead of electronic health record (EHR) systems, which take up less time for real human interaction. In less-resourced languages such as Finnish, in which natural language processing (NLP) tools are only beginning to emerge, this is an even bigger challenge. This study investigates the fine-tuning of the open-source LLaMA 3.1-8B language model on simulated Finnish clinical conversations that is, transcribed clinical dialogues created by Metropolia UAS students. The aim is to verify if a domain aligned large language model (LLM) is able to reliably translate spoken Finnish medical discourse into formal transcript document when it train with a small dataset. With 7-fold cross-validation, the fine-tuned model achieved a BLEU score of 0.1214, ROUGE-L score of 0.4982, and BERTScore F1 score of 0.8230, showing satisfactory semantic performance using a small dataset and scalability of privacy-oriented NLP tools in Finnish medical environments.
\end{abstract}

\begin{IEEEkeywords}
LLM, Medical Transcription, NLP, Finnish Language, LLaMA.
\end{IEEEkeywords}

\section{Introduction}
In a modern healthcare system, clinical documentation is absolutely essential for patient safety, correct diagnosis, support of medical billing, and preservation of legal documents. However, the complexity in the health care system and the complexity in the patients' conditions have increased the burden on the health care professionals in documenting the patients' records, which has led to a decrease in the time spent interacting with patients, leading to burnout among physicians \cite{10.1001/jamanetworkopen.2024.26956}. This burden is mainly on the doctors and nurses, who have to divide their time between interacting with the patients and documenting the patients' records in electronic health records. This has a negative impact on the quality of interaction with the patients and the quality of the patients' records. This not only affects the quality of service provided by the health care professionals but also impacts the health care professionals' satisfaction and well-being. It has also led to confusion in the patients' records, which has a negative impact on the health care services provided, especially in complex patient conditions.
% In a modern healthcare system, clinical documentation is absolutely essential for
% patient safety, correct diagnosis, support of medical billing, and preservation of
% legal documents. The increasing complexity of patient interactions and
% healthcare delivery systems has resulted in a greater burden of documentation
% for clinicians, which frequently causes physician burnout and a decrease patient
% interactions time and structural clinical documentation helps doctors and nurses
% to give appropriate treatment to patient and follow up. Traditionally, this
% responsibility has fallen heavily on physicians and nurses, contributing to widespread documentation burden and professional burnout. Physicians frequently have to split their time between talking to patients and entering information into electronic health records (EHRs), which can compromise both the depth of patient communication and the completeness of clinical documentation. This cognitive stress lowers the quality of care provided while also raising the risk of burnout and decreasing work satisfaction. On the other hand, patients may have recurrent questions, disjointed treatment pathways, and impeded continuity between experts as a result of inconsistent or poorly organised data, especially in complicated or chronic care cases.
\begin{figure}[htbp] 
\centering 
\includegraphics[width=\linewidth]{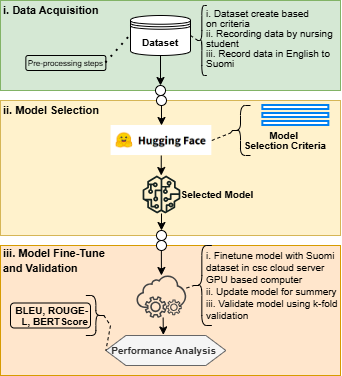} \caption{Overview of Fine-Tuned LLM model using Small Validate Dataset.} 
\label{fig:architecture} 
\end{figure}

Growing concerns among healthcare stakeholders have emphasized that the increasing volume of clinical documentation directly affects the quality of interactions between patients and clinicians in consultations. These issues, in turn, highlight the need to adopt a structured approach to clinical documentation, in which patient data is systematically organized in a structured, normalized, and easily understandable format. This approach to structuring clinical data promotes greater interoperability between healthcare systems, facilitates more informed clinical decision-making, and enables secondary uses of data, such as population health management or medical research \cite{Sasseville2025Impact}. In addition, this approach ensures that vital clinical information, such as diagnoses, medications, allergies, or follow-up recommendations, is recorded in an accurate, accessible, and consistent manner to enhance the quality of patient care.

% Beside that, all of the stakeholders are becoming increasingly concerned about
% the burden of clinical paperwork since it has a direct effect on the calibre of
% interactions during consultations. These challenges underscore the critical need
% for structured clinical documentation, where patient information is organized into
% standardised, human-readable formats. Structured documentation facilitates
% interoperability across health systems, supports evidence-based clinical
% decision-making, and enables secondary uses such as population health
% analytics and health services research. It makes ensuring that important clinical
% observations are regularly, accurately, and retrievably recorded, including
% diagnoses, prescriptions, allergies, and instructions for follow-up, thereby
% improving care outcomes and long-term patient safety.\\

The incorporation of medical scribes, whether human or digital, has surfaced as an effective strategy to alleviate administrative burdens and enhance workflow efficiency \cite{Miksanek2020Productivity}. Human scribes assist physicians by documenting patient encounters in real time, allowing clinicians to focus more on patient interaction rather than data entry into electronic health records. While this approach has been shown to enhance workflow and increase physician satisfaction, it also presents challenges such as high costs, limited scalability, potential inconsistencies in documentation quality, and concerns related to data privacy and standardized training.

The limitations of human scribes have led to the rapid adoption of AI \cite{Mess2025Artificial}, especially in the form of automatic speech recognition and natural language processing (NLP), to facilitate the development of digital medical transcription tools. Recent developments in large language models (LLMs) have also led to the significant improvement of the ability to process and organize complex language \cite{Hassan2025Clinical}. In terms of the current study, which assesses the effectiveness of a fine-tuned LLM for Finnish medical transcription using a validated dataset, the open-source nature of LLMs provides a significant advantage, especially in terms of the ability to adapt the model to the needs of the specific domain while ensuring the privacy of the data. Fine-tuning the LLM on a limited but high-quality dataset of Finnish clinical language can facilitate the effective transformation of unstructured speech into standardized medical documentation.

However, in non-English languages, especially those with rich morphological traits like Finnish, the construction of such systems becomes much more challenging. Finnish (Suomi) is a morphologically agglutinative language, indicating that a single word can represent several grammatical characteristics, leading to a highly sparse vocabulary and complex parsing challenges. In the medical domain, this complexity is compounded by specialized terminology, code-switching (e.g., Latin, English loanwords), and the need for high accuracy due to patient safety concerns \cite{H2021Current}.

Open-source LLMs represent a promising approach in the context of a health environment with strict privacy regulations such as GDPR in place because of the benefits of transparency and customization options along with the opportunity of local deployment in a secure manner. These models have the advantage of being adapted to specific domains using specific datasets in specific languages such as Finnish, which allows them to recognize complex linguistic features such as morphology and medical terminology. Therefore, this is in line with the objectives of this specific study, which is focused on the evaluation of a fine-tuned LLM in the context of Finnish medical transcription using a small validated dataset. This is especially relevant in the context of the development of accurate and privacy-compliant transcription systems. There is a great opportunity of minimizing documentation work with the help of this approach in order to provide better care with the assistance of such a system. Consequently, this study proposes an advance approach: fine-tune open-source LLMs to accurately the transformation of transcribed clinical speech into structured documentation, specifically for the Finnish language. Fig.~\ref{fig:architecture} shown the architecture of Evaluating Fine-Tuned LLM model using Small Validate Dataset of low-resourced languages such as Finnish (Suomi).

\section{Literature Review}

Healthcare's digital transformation has placed increasing pressure on the efficiency, accuracy, and structure of clinical documentation \cite{Adedeji2024Sound}. As physicians face mounting administrative demands, accurate record-keeping has become both a critical necessity and a major burden. Clinical documentation is essential to quality assurance, billing, diagnosis, treatment continuity, patient safety, and compliance with rules. However, because of the enormous time commitment needed, many doctors are now more burned out and dissatisfied, spending more time documenting than connecting with patients.

To minimize the documentation time, the use of human medical scribes has been advocated, who assist in recording the clinical encounters in real time, thus allowing the physician to concentrate more on the patients \cite{Sasseville2025Impact} \cite{Mess2025Artificial}. However, the use of medical scribes has also been found to have some drawbacks, such as the need to incur heavy training costs, the quality of the documentation process, and the issue of privacy, which has shown the need to explore more effective alternatives, such as the fine-tuned LLM-based Finnish medical transcription system, as researched in this study.

Limitation of human scribes has led to the development of AI-based digital scribe systems, which utilize ASR and NLP techniques for clinical documentation generation \cite{Adedeji2024Sound}. The performance of ASR has improved significantly in recent times, especially for open-source tools such as GPT, DeepSpeech, and Whisper, which utilize deep learning techniques for speech recognition. This study, therefore, seeks to assess the performance of a fine-tuned LLM on a small validated dataset for the improvement of speech recognition in the context of Finnish medical transcription, which may be affected by domain-specific terminologies, linguistic complexity, and code-switching in noisy backgrounds.

After transcription, unstructured clinical conversations must be converted into structured formats to support decision-making and system interoperability. Recent advancements in open-source LLMs have demonstrated promising results in extracting key medical information and producing structured outcomes from unstructured dialogues \cite{Garcia2024Medical}. In this particular study, a fine-tuned LLM is assessed for medical transcription in Finnish, using a small but validated dataset, based on its capacity to transform unstructured dialogues into structured medical documents.

Complexity is a fundamental attribute of clinical conversations, which are often interrupted, ambiguous, and colloquial in nature, making transcription and structuring more difficult compared to regular medical records \cite{Khatim2024Using}. However, recent breakthroughs in the field have demonstrated the ability of fine-tuned LLMs to successfully transcribe such conversations, which are typically represented in a multi-turn dialogue format, outperforming regular NLP approaches. In this study, a fine-tuned LLM is tested in the field of Finnish medical transcription, using a small validated data set.

While recent studies have shown some positive results in the area of medical transcription and LLMs, most of them have focused on the English language, with little work being conducted in lesser-resourced languages such as Finnish \cite{Yang2025Fine}. The language of Finnish, with its complex morphology and word order flexibility, adds to the complexity of the task, especially considering the lack of resources available for the task. While tools such as FinBERT, OPUS-MT, and AaltoASR are useful, domain-specific models need to be fine-tuned. The present work aims to assess a fine-tuned model for Finnish medical transcription with a small validated data set.

This study is based on the concept of transfer learning \cite{Vrbancic2020Transfer}, where language models are fine-tuned on specific domain and language data to enhance performance. Previous studies have shown that domain-specific language models perform better than general ones in clinical tasks. This study is based on the concept, and it aims to test whether a fine-tuned LLM can produce accurate and structured Finnish medical transcription using a small validated dataset.

\section{Methodological Approach}

This section outlines the process for data creation for fine-tuning, selecting open-source LLMs, pre-train model finding from hugging face platform, and defining the evaluation methodology. 

\subsection{Data Creation}\label{sec:Data_Creation}
The data creation process began with the establishment of a foundational dataset comprising Finnish clinical case conversations. These baseline conversations were formatted simulated interactions between healthcare professionals such as doctors or nurses and patients. However, students from the Innovation Project \cite{metropolia_innovation_2025}, play those roles in creating this dataset (cf. Step-I, in Fig.~\ref{fig:architecture} ). The aim was to capture authentic, context-rich dialogue reflective of real-world clinical scenarios. This baseline serves as the core reference for further development and training of conversational models or language-processing systems in the Finnish healthcare context.

To build the dataset, each clinical scenario was documented in two formats: an audio recording in MP3 format and a corresponding textual transcription. The MP3 files represent the actual spoken dialogues, while the text files offer human readable versions of these recordings in Finnish. These transcriptions are aligned with the audio to ensure accuracy and preserve the nuances of clinical communication, including terminology, emotional tone, and conversational flow.

All records in the dataset follow the same naming pattern for traceability and for structural organisation. Files are encoded in the format I01-G01-C01.txt and I01-G01-C01.mp3, where each component possesses some metadata. The prefix I01 indicates iteration number of the batch of the dataset. The middle segment G01 delineates the specific group of students in the innovation projects who contributed to the creation of that data entry, facilitating the attribution of performance and contribution between groups of students. The final segment C01 is the clinical case number, which links the file to an individual patient scenario. This naming pattern is applied uniformly to both audio (MP3) and text (TXT) files to facilitate simple management, retrieval, and referencing within research, analysis, or model training pipelines.

Students from different disciplines, like nursing, podiatry, paramedic, and gerontology, worked out to carry out the production process, and this process was monitored by their teachers. Under supervision, these students actively contributed to the creation and transcription of the clinical dialogues, guaranteeing clinical authenticity and realism. Their multidisciplinary contribution was critical in the creation of a dataset that achieves a balance between healthcare usefulness and pedagogical usefulness. The generated resource supports instructional use cases in nursing and healthcare simulation settings and offers a basis for training AI systems in clinical communication in the Finnish language.

\subsection{Hugging Face Platform}\label{A}
The development of the so-called Transformer architectures signaled a major breakthrough in the field of NLP, allowing for a level of scalability, efficiency, and effectiveness beyond the RNN-based architectures \cite{Wu2025LLM}. Building upon the effectiveness of the pre-training of language models on large datasets, the study aims to fine-tune the LLMs for the task of medical transcription in Finnish, a task that requires efficiency and effectiveness.

As NLP models became more advanced, the need for accessible and adaptable tools led to the rise of platforms like Hugging Face, which support open-source model sharing and fine-tuning \cite{Pol2024Hugging}.  These platforms allow researchers to easily fine-tune pre-trained models for specific tasks. In the study, these tools are used to fine-tune LLMs for medical transcription in Finnish using a small set of validated data.

Hugging Face is a popular platform in the machine learning community with its open-source ecosystem providing a variety of models of the Transformer type, which can be easily fine-tuned and deployed with the help of its tools \cite{Ait2023HFCommunity}. The platform is highly suitable to be used to adapt models to a particular domain or application. In the present study, the platform is used to create and test a fine-tuned LLM model for the application of medical transcription in the Finnish language with the help of a small validated set of data.

In fact, Hugging Face offers a wide range of tools for developing domain-specific language models through its integrated tools for datasets, pre-trained models, as well as efficient deployment of language models in a collaborative environment \cite{Susnjak2025Automating}. This study utilizes the tools provided by Hugging Face to develop and evaluate a fine-tuned LLM model for efficient transcription of medical documents in Finnish using a small validated dataset. Step-II, in Fig.~\ref{fig:architecture} the approach how Hugging Face used in this research.

In addition, Hugging Face contributes to both academic and practical research by supporting open-source development, making it easy to reproduce results, and providing easy access to state-of-the-art tools in NLP research \cite{Shen2023HuggingGPT}. The tools provided by Hugging Face, including efficient tools for fine-tuning models and deploying models, allow researchers to fine-tune pre-trained models according to specific domains or languages. This research utilizes the tools provided by Hugging Face to fine-tune an LLM model for medical transcription in the Finnish language using a small validated dataset.

\subsection{Selection of LLM Model}\label{B}
For this research, selecting the appropriate Large Language Models (LLMs) was guided by a set of criteria that align with the specific requirements of structured clinical documentation in Finnish. The goal was to identify models that not only demonstrate a high level of accuracy in medical language processing but also provide flexibility for customisation and support for the Finnish language. The core selection criteria included: 

\paragraph{Ability to handle medical text optimisation} Large language model parameters refer to the model’s internal variables, which are the weights and biases learned during the training process. Parameters influence the effectiveness of the model in identifying linguistic patterns, sensing the context, and generating meaningful language \cite{Zhiqiang2023LLM}. Parameters’ size and configuration influence the complexity and effectiveness of the model.

\paragraph{Support for structured clinical documentation} Structured clinical documentation is the process of taking unstructured medical information such as free-text conversation or handwritten orders and placing it into a standardized, structured format \cite{Sasseville2025Impact}. It enables quicker data retrieval, clinical decision making, health system interoperability, and secondary use such as analytics and research.

\paragraph{Customisation and fine-tuning capabilities} 
A pre-trained language model for the Finnish language is first trained on a large dataset of Finnish text to acquire language patterns, including grammar and vocabulary, and can be further fine-tuned on a specific dataset for specific tasks \cite{Lankford2023adaptMLLM}. In the study, the method is used to fine-tune an LLM model for medical transcription in the Finnish language using a small validated dataset.

\paragraph{Finnish language support, pre-training or adaptability for continued learning} An adaptable model can be modified or extended to accommodate specific research objectives or domain requirements \cite{Chen2024Meta}. It includes concepts like modification of the model structure, addition of new data sets, or tuning of the training objectives for optimal task-specific performance.

Although models like BioGPT and ClinicalBERT have proven efficiency in medical-related tasks, they do not have the Finnish language option. However, models like FinBERT, which have been designed for the Finnish language, have proven efficiency \cite{Maity2025Large}. In this study, appropriate open-source LLMs have been identified and adapted for the purpose of balancing efficiency and flexibility in the transcription of medical texts in the Finnish language using a small validated dataset. Based on all criteria and models’ availability on Hugging face we developed a models comparison table to solve this selection stage, Table~\ref{tab:model_comparison} explain it in detail.

\begin{table}[htbp]
\caption{Models' comparison based on selection criteria}
\label{tab:model_comparison}

% Make table text larger
\large

% Left align the whole table
\begin{flushleft}

% Resize to fit column but keep readable
\resizebox{1\columnwidth}{!}{%
\begin{tabular}{|p{3.5cm}|p{2.2cm}|p{2.5cm}|p{2.5cm}|p{2.4cm}|}
\hline
\textbf{Model Name} & 
\textbf{Parameters (B/M)} & 
\textbf{Structured Clinical Documentation} &
\textbf{Finnish (Pre-trained / Fine-tune)} &
\textbf{Customisable} \\
\hline
Llama 3 (Meta) \cite{Hugo2023LLaMA} & 8B, 70B, 405B & Yes & Yes & Yes \\
\hline
SiloGen Finnish GPT \cite{amd2023finnishai} & 7B, 13B, 33B  & Yes & Yes & Yes \\
\hline
Teuken-7B (OpenGPT-X) \cite{Mehdi2024Teuken} & 7B & Yes (with fine-tuning) & Yes & Yes \\
\hline
Hippocrates LLM \cite{Acikgoz2024Hippocrates} & 7B, 13B  & Yes & No & Yes \\
\hline
DeepSeek  \cite{2024DeepSeek} & 7B, 67B & Yes (with fine-tuning) & Yes & Yes \\
\hline
TurkuNLP FinBERT \cite{virtanen2019multilingual} & 110M          & Yes (with fine-tuning) & Yes & Yes \\
\hline
BioGPT (Microsoft) \cite{Luo2022BioGPT} & 1.2B  & Yes & Yes & Yes \\
\hline
ClinicalBERT  \cite{Matondora2024NLP} & 110M & Yes & No & Yes \\
\hline
\multicolumn{5}{l}{\textit{All models support customization through fine-tuning or prompt engineering.}} \\
\hline
\end{tabular}
}
\end{flushleft}
\end{table}

Based on the evaluation criteria, the model chosen by the study is LLaMA 3.1-8B, as it has proven its efficiency, scalability, and adaptability. Being an open-source LLM model, it allows for efficient fine-tuning methods, making it more suitable for domain-specific applications, including Finnish medical transcription. The model is also accessible via platforms such as Hugging Face, making it more viable for use in the study. As a result, the model is fine-tuned using a small validated data set in order to produce accurate documentation in Finnish.

\subsection{Evaluation Methodology}\label{C}
This section of the study gives a clear explanation of the method that was used to assess the performance of the fine-tuned model. Specifically, it outlines how model outputs were contrasted with reference clinical notes made by humans for the purposes of determining the quality of structured documentation that is generated from clinical conversations in the Finnish language.

In the proposed study, the quality of Finnish medical transcriptions produced by the fine-tuned LLM will be assessed using the BLEU, ROUGE, and BERTScore methods \cite{Tianyi2022BERTScore}. The proposed methods will allow the assessment of the medical transcriptions' quality by comparing them to validated medical references.

The BLEU score is used to compare the lexical similarity of the Finnish medical transcription generated with the reference clinical notes through the evaluation of the precision of n-gram \cite{Glushkova2023BLEU}. It is also used to determine the syntactical accuracy of the clinical terms used in fine-tuned LLM.

The ROUGE metric, which includes ROUGE-N and ROUGE-L, is used to test the quality of the Finnish medical transcription texts produced by the fine-tuned LLM. This is done by comparing the texts with the reference clinical notes. The ROUGE-N metric works using n-grams, whereas the ROUGE-L metric works by finding the longest subsequence using the longest common subsequence algorithm \cite{Barbella2022Rouge}. This is important in medical contexts because it determines whether important medical details have been accurately captured despite the variation in the wording.

 To assess the semantic similarity between the produced and reference material, BERTScore is included to go beyond surface-level lexical matching. Unlike the word-based comparison with BLEU or ROUGE, BERTScore uses contextual word representations to compare the meaning of the output with the target output \cite{Tang2024Improving}. In the study, BERTScore is integrated to compare the Finnish medical transcription output generated with the LLM with the target output, thus making the model effective in capturing clinically equivalent expressions.

These three indicators work together to offer a comprehensive framework for evaluation. ROUGE gauges content recall, BERTScore records semantic integrity, and BLEU quantitatively assesses syntactic accuracy. This complete evaluation technique facilitates an impartial assessment of the model's capacity to produce accurate, therapeutically pertinent, and cohesive documentation from Finnish clinical dialogues.

\subsection{K-Fold Cross-Validation}\label{D}
Cross-validation is a popular method used to test the generalizability of models in machine learning algorithms, especially when working with small datasets. In the present study, the model uses the K-fold cross-validation method because of the small size of the validated Finnish medical transcription dataset \cite{Department2021Performance}. In the K-fold cross-validation method, the data is divided into K-folds, and the model is trained on all the data to ensure the efficient use of the data to test the model's performance.

\begin{figure}[htbp] 
\centering 
\includegraphics[width=\linewidth]{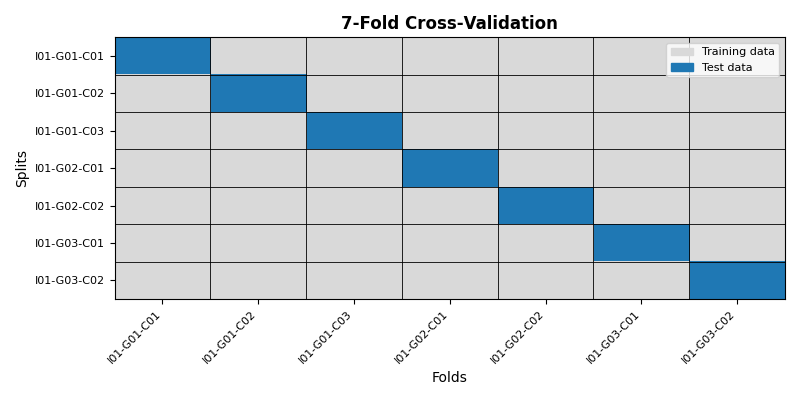} \caption{K-Fold Cross-Validation (k=7) Structure.} 
\label{fig:kflod_validation} 
\end{figure}

In this study, the data set is limited to only seven clinical dialogues in Finnish with their corresponding MP3 files and manually annotated reference documents. Due to the limited number of samples, standard single-split validation would lead to unstable and biased performance estimates. To combat this limitation, a 7-fold cross-validation (leave-one-out) method is adopted where a single conversation is employed as test case while learning is conducted on the remaining six, which is dispatch in the fig~\ref{fig:kflod_validation}. This allows for a greater overall understanding of the model's behaviour under various patient interactions and clinical subjects, which is essential in the healthcare sector where data diversity is as essential as data volume.

K-fold cross-validation is advantageous when using LLMs to fine-tune domain-specific models since it reduces the variance due to random sampling, guarantees the stability of model performance with different samples, and detects any inconsistencies in the dataset \cite{Lin2022Curriculum}. For a low-resource domain like Finnish medical transcription, the method is advantageous when avoiding overfitting is a priority through multiple tests of model generalizability. Together with BLEU, ROUGE, and BERTScore metrics, the model can be evaluated syntactically and semantically to ensure the scientific rigor of the model’s performance assessment.

\section{Experimental Study and Analysis}
This section explains the experiment configuration, dataset preparation, fine-tuning process, and evaluation method of the study. It discusses the test of the fine-tuned LLM model on Finnish clinical conversations, including the results obtained as well as analysis based on automatic metrics.
\subsection{Dataset Summary}\label{E}
As explained in Section~\ref{sec:Data_Creation}, the study utilized a custom dataset drawn from Finnish-language clinical MP3 records and their manually written textual equivalents. The dataset was specifically curated to reflect clinical conversation in Finnish, capturing a variety of medical scenarios and communication styles. Audio recordings were generated from existing text data to ensure that the resulting transcriptions maintained both clinical appropriateness and linguistic precision, supporting effective natural language processing.

The dataset is comprised of 7 complete clinical dialogues. Pre-processing was applied to each sample and transferred into structured JSON format specially to cater to the training requirements of large language models. The JSON format has a clear segmentation of dialogue turns, speaker designation (e.g.,doctor or patient), and annotated fields that indicate the target output for clinical documentation. This format allowed for supervised fine-tuning, which allowed the model to learn how to convert free-form clinical discussion into well-formed, documentation that met the standards of healthcare.

\subsection{Experimental Setup}\label{F}
In this section, the technical setup for model training and evaluation is explained, which was carried out on the CSC Puhti supercomputer to provide a secure environment for fine-tuning the large language model \cite{csc_puhti}. The setup included hardware with GPU optimization along with necessary ML libraries and frameworks to support the training of the model, particularly the transformer-based model. The setup was effective for the adaptation of the LLM.

\paragraph{Software}
This work uses Python Ver. 3.11.5, PyTorch version $\geq$ 2.0 with Compute Unified Device Architecture (CUDA) 11.7, and open source ML libraries on a GPU based Puhti supercomputer for fine-tuning, validation, and quantitative evaluation \cite{python3115} \cite{cuda117}.

The project leveraged basic machine learning library tools, PyTorch for model training, and CUDA for parallel processing via the supercomputer Puhti. The pre-trained LLaMA model and tokenizer are provided by the Hugging Face library, while the datasets library was used for efficient data management. Clinical voice recordings in Finnish language are initially transcribed using the Whisper LargeV3 model, followed by pre-processing in a standardized manner for fine-tuning the LLM model.

\paragraph{Hardware}
The fine-tuning experiments were conducted on a CSC Puhti supercomputer with one NVIDIA A100 GPU and 40 GB VRAM to fine-tune LLaMA 3.1-8B in 4-bit quantized form with LoRA adapters. The system had 16 CPU cores and 128 GB system memory to accommodate data preprocessing, parallel task execution, and model loading. Additionally, 200 GB disk space was available to store model weights, outputs, and data. This setup was a good compromise between computational efficiency and model demands to enable fine-tuning and evaluating models for Finnish medical transcription with stability and speed.

\subsection{Experimental Setup on CSC Puhti Supercomputer}\label{G}
To begin the final experimental setup, an account was created on the CSC cloud service, then a project was created based on the title of the study. After approval of the project and receiving the project ID, as a student of Metropolia University of Applied Sciences, we were given 100,000 billing units to have full access to the Puhti supercomputer and CSC cloud services to train, tune, and test all the models without any cost for the Finnish medical transcription service.

The project was executed on a GPU node that satisfied all hardware and software requirements, including a single NVIDIA A100 GPU with 40 GB of VRAM, 16 Xeon CPU cores, 128 GB of RAM, and approximately 200 GB of local storage. This provided a balance of memory and computational resources required for fine-tuning the 8-billion-parameter LLaMA 3.1 model in 4-bit quantized form with the use of LoRA adapters. The CPU cores also enabled the parallel processing of data preprocessing and I/O. Additionally, the high-speed interconnect of the Puhti cluster, along with the Lustre file system, provides a reliable level of reproducibility in the performance of the project. Before coding the project, the research team obtained the necessary access to the LLaMA 3.1-8B model through the Hugging Face library. This was done in accordance with the terms of the license, the responsible use of the model, and the ethical guidelines established by the organization. Once the necessary approval was granted, the model's weights were downloaded into the environment, allowing for the fine-tuning of the model for the purposes of Finnish medical transcription.
% Before beginning the coding and tuning phases of the project, obtaining access to the LLaMA 3.1–8B model on Hugging Face through Meta's official repository was required. Since the LLaMA models are released under a bespoke license with some restrictions regarding how they may be utilized, access is not granted by default. To proceed, the research team were required to request access to the models by submitting an application through the Hugging Face platform. This involved consenting to Meta's terms of license, which outline permissible use cases, limitations, and citation policies. The applicants were also required to sign off on the responsible use form, confirming compliance with legal and ethical standards, including non-commercial use and safe deployment. Once approved, model weights were downloadable and could be included in the project environment. This ensured all model usage was done under terms described by Meta to enable proper and secure deployment of LLaMA 3.1–8B in the fine-tuning pipeline.

\subsection{Finetuning}\label{H}
The fine-tuning of the LLaMA 3.1-8B model marked a significant milestone in the development of a specialized LLM from a general-purpose LLM, with the broader objective of fulfilling healthcare-related educational objectives. The training data comprised MP3 file formats of clinical conversations, along with their corresponding text-based versions, wherein students of the Innovation Project, act healthcare professional (e.g., doctor, nurse, therapist) \cite{metropolia_innovation_2025}, and patient are the roles played in the development of the dataset. In the next step, the audio recordings were transcribed using the Whisper large-v3 model. Subsequently, the audio-transcript data was pre-processed into a standard format for the development of the LLM. Five of the audio-transcript files were chosen as the initial training and test files for two files each. In the subsequent step, a 7-fold cross-validation method was adopted for the development of the LLaMA 3.1-8B model. This method helped the model learn from the dataset repeatedly, thereby avoiding the chances of overfitting. This cross-validation method also helped the model attain higher reliability for its application in the development of clinical documentation and conversational AI.

The fine-tuning was conducted using the \textit{meta-llama/Llama-3.1–8B} model, accessed from a local directory after obtaining the necessary license from Meta. The model was added to memory with \textit{float16} precision and \textit{device-map="auto"} to utilize available GPU memory capacity on the CSC Puhti supercomputer. The LLaMA architecture-compatible tokenizer was also initialized with the original tokenizer model and utilized to convert textual training data into sequences of tokens. Records were padded or truncated up to a maximum of \textit{512-tokens} to maintain input lengths consistent during training. This \textit{512-tokens} was chosen as the input length limit to maintain transformer structure compatibility, with tractable memory usage in training while still gathering adequate clinical context.

The training involved using Hugging Face's Trainer API in half-precision supervised fine-tuning mode. Significant hyperparameters included a batch size of one (1), three (3) epochs of training, logging at every \textit{10-steps} (e g. 6 training samples × 3 epochs = 18 steps per fold), and saving the checkpoint at every \textit{100-steps} with a limit of 2 saved checkpoints. Gradient check pointing was not used
under this setup because the training setup was made minimal for the sake of clarity and stability. The models were trained in half-precision floating point \textit{(fp16=True)} to maintain the calculation below the memory constraint of the available NVIDIA V100 GPUs on Puhti.

The training script included a custom tokenizer and data collator for causal language modelling. The \textit{DataCollatorForLanguageModeling} was used with masked language modelling disabled \textit{mlm=False)}, aligning with the \textit{LLaMA} model's causal architecture. No evaluation was conducted during training \textit{(evaluation-strategy="no")}, as the focus was on completing a robust initial finetuning pass on the full training set.

Following the process of training, the fine-tuned model was saved in different forms. First, the model and tokenizer were saved in Hugging Face transformers format through \textit{save-pretrained()} for convenient reloading as well as continued development. Additionally, the PyTorch state dictionary for the model was saved in \textit{.pth} format for non-Hugging Face out-of-the-box applications or \textit{non-Hugging Face} frameworks compatibility. While \textit{LLaMA} models themselves aren't natively \textit{TensorFlow/Keras} compatible, there was a placeholder \textit{.h5} file included to show potential future need for a conversion, though full \textit{.h5} export is not currently supported.

All computations were done with \textit{SLURM job} scripts on the Puhti GPU cluster, with GPU reservations requested by \textit{--gres=gpu:v100:1}, \textit{16 CPU cores}, and \textit{128 GB RAM}. The virtual Python environment was also activated within the job script to assure stable dependency management and reproducibility.

This fine-tuning process produced a domain-tuned \textit{LLaMA-3.1–8B} model with the ability to generate Finnish clinical conversation texts in the given contexts. The model can now translate spoken Finnish medical discourse between patients and healthcare professionals into neatly organized clinical conversation documents. Trained model and its training and preprocessing scripts are made available publicly at \textit{GitHub} facilitating future research and deployment in clinical AI tools.

\section{Results and Evaluation}

To evaluate the model ability after fine-tuned, research conducted automatic testing using \textit{BLEU$, $ROUGE-L}, and \textit{BERTScore} measures. These analyses were applied between generated documents and ground-true documents of Finnish clinical conversations. The evaluation process took a $7-fold$ cross-validation method to ensure proper performance estimation on different subsets of the data and prevent overfitting with respect to the finite amount of data available, Table~\ref{tab:results} illustrated the results of \textit{k-fold} validation. This helped in ensuring stringent evaluation of the model's capability to generalize to clinical conversations generation and documentation tasks.

\begin{table}[htbp]
\caption{7-Fold Cross-Validation Results}
\label{tab:results}
% Make table text larger
\large
% Left align the whole table
\begin{flushleft}
% Resize to fit column but keep readable
\resizebox{1\columnwidth}{!}{%
\begin{tabular}{|p{3.5cm}|p{3cm}|p{3cm}|p{3cm}|}
\hline
\textbf{File Names} & \textbf{BLEU} & \textbf{ROUGE-L} & \textbf{BERTScore} \\
\hline
I01-G01-C01.txt &  0.0739 & 0.3156 & 0.7642 \\
\hline
I01-G01-C02.txt & 0.1579 & 0.5151 & 0.8384 \\
\hline
I01-G01-C03.txt & 0.1154 & 0.5759 & 0.8621 \\
\hline
I01-G02-C01.txt & 0.1330 & 0.5829 & 0.8472 \\
\hline
I01-G02-C02.txt & 0.1247 & 0.4558 &  0.7905\\
\hline
I01-G03-C01.txt & 0.1151 & 0.5447 & 0.8390 \\
\hline
I01-G03-C02.txt & 0.1295 & 0.4974 & 0.8197 \\
\hline
\textbf{Average} & \textbf{0.1214} & \textbf{0.4982} & \textbf{0.8230} \\
\hline
\multicolumn{4}{l}{\textit{All data files used in k-fold validation.}} \\
\end{tabular}
}
\end{flushleft}
\end{table}

The $BLEU$ score was employed to evaluate the precision of n-gram overlaps between the fine-tuned LLM model output and reference Finnish clinical transcripts, which reflected the similarity between the model output and actual human conversations. The \textit{ROUGE-L} score, with a focus on recall, was employed to evaluate the completeness of the generated text with respect to the reference text. Moreover, \textit{BERTScore} was employed to evaluate semantic similarity, which reflected the similarity in meaning, a common phenomenon in clinical contexts.
% BLEU was employed to evaluate the precision of n-gram overlaps between the model's outputs and reference transcripts. It served as a key indicator of how well the model replicated human-authored clinical conversation. $ROUGE-L$ emphasizing recall, was used to assess how comprehensively the generated text matched the reference content. Additionally, we included $BERTScore$, a semantic similarity metric based on contextual embeddings, to better capture the meaning-level fidelity of the outputs, especially important in the healthcare context where paraphrasing and variation in wording are common.

\begin{figure}[htbp] 
\centering 
\includegraphics[width=\linewidth]{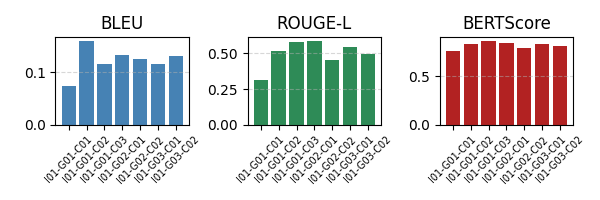} \caption{Model results in k-fold validation.} 
\label{fig:result_1} 
\end{figure}

Across \textit{7-fold} cross-validation, the model achieved a mean \textit{BLEU} score of \textit{0.1214}, \textit{ROUGE-L} score of \textit{0.4982}, and \textit{BERTScore} of \textit{0.8230}. These all reflect moderate syntactic similarity and high semantic matching with human-transcribed references due to the complexity and domain-specific nature of the input data. In general, 10–20 \% \textit{BLEU} scores are generally moderate, particularly in low resource and high-variety settings. \textit{ROUGE-L} scores over $0.5$ indicate adequate overlap between generated and reference text. By contrast, \textit{BERTScore} scores of above $0.8$ represent high semantic similarity, with scores of $0.85$ or higher being generally regarded as representative of high-quality semantic-level correspondence. These test results confirm that fine-tune LLMs model, \textit{LLaMA-3.1–8B}'s capability to generate contextually appropriate and linguistically coherent outputs from Finnish clinical audio inputs. The 7-fold cross-validation complemented the credibility of the performance metrics by demonstrating stable outcomes across diverse data partitions. In fig~\ref{fig:result_1} illustrated the results of all dataset based on \textit{BLUE}, \textit{ROUGE-L}, and \textit{BERTScore}. In conclusion, the evaluation supports the model’s readiness for downstream applications in Finnish-language clinical NLP systems.

\section{Discussion and Conclusions}
The fine-tuning of the \textit{LLaMA-3.1-8B} model for generating Finnish clinical conversation documents illustrates the utility of domain-specific fine-tuning of LLMs. Through the use of hand-curated transcripts and Whisper transcriptions, the model was able to demonstrate strong capabilities in understanding and generating clinical content in Finnish. Through evaluation metrics, it is evident that the fine-tuned model was able to successfully capture both syntactic accuracy and semantic coherence, even with the limited dataset. This illustrates the utility of specialized LLMs for tasks such as clinical documentation and education in low-resource languages.

The fine-tuned \textit{LLaMA-3.1–8B} model was more in line with the linguistic nuances of Finnish healthcare communication. While larger models tend to outperform smaller models on general benchmarks due to large-scale pretraining, this study demonstrates that task-specific fine-tuning even on a limited dataset.

The k-fold cross-validation improved the model's strength and reduced overfitting issues, especially with domain-specific data like clinical texts. It improved the generalizability of the model towards diverse input patterns. While \textit{BERTScore} (0.8230) reported high semantic retention, \textit{BLEU} (0.1214) and \textit{ROUGE} (0.4982) suggested there were areas to be worked on in precision and recall. Using training on the CSC Puhti supercomputer with \textit{fp16} optimization and \textit{LoRA} validated the process to be sustainable using limited resources, albeit parameter fine-tuning is required for replicating the process in less-capable machines.

A key limitation of this study is the small size of the dataset, which consisted of only seven Finnish clinical dialogues, and the fact that the \textit{LLaMA-3.1–8B} model which is the only model used, is not natively pre-trained in Finnish. These constraints may affect the model's ability to learn linguistic nuances as well as clinical domain-specific patterns. Future work could involve expanding the training dataset with more authentic or synthetic Finnish clinical conversation to further enhance model generalization, and there should be train others models and comparison between the models, the baseline results (without pre-training) will also need to evaluate. Moreover, additional architectural refinements, such as incorporating structured end-of-sequence tokens or leveraging retrieval augmented generation, may increase output accuracy and reduce hallucination. As shown in related works using models like \textit{TurkuNLP}, \textit{FinBERT}, or \textit{ClinicalBERT}, smaller LLMs can be made more effective through such augmentation, suggesting potential for lighter deployment scenarios in healthcare systems with resource limitations.

In the end, while the study was based on automatic evaluation, it shows the possibility of using human-in-the-loop feedback in the future. Healthcare professionals and Finnish language experts could be used as a validation tool, which could give better results in terms of the quality of the output of the model. The results of the study show that fine-tuned LLMs, as in the case of \textit{LLaMA-3.1-8B}, could be a good basis in the development of intelligent medical scribes and conversational AI in Finnish settings.

% \subsection{Equations}
% Number equations consecutively. To make your 
% equations more compact, you may use the solidus (~/~), the exp function, or 
% appropriate exponents. Italicize Roman symbols for quantities and variables, 
% but not Greek symbols. Use a long dash rather than a hyphen for a minus 
% sign. Punctuate equations with commas or periods when they are part of a 
% sentence, as in:
% \begin{equation}
% a+b=\gamma\label{eq}
% \end{equation}

\section*{Acknowledgment}
The authors would like to thank Digital Medical Scribe project team members, Principal Lecturer Mikael Soini, Principal Lecturer Päivi Haho, Senior Lecturer Sakari Lukkarinen and supervisors of Nursing, Bachelor's Degree, Nursing Bilingual Top-Up Degree, Biomedical Laboratory Science, Bachelor's Degree, Laboratory Science, Bachelor's Degree, Bachelor of Health Care, Physiotherapist (top-up), and Bachelor of Health Care, Physiotherapist for their support in collecting the dataset, and direct and indirect contribution to the project.

\bibliographystyle{IEEEtran}
\bibliography{reference}

@article{10.1001/jamanetworkopen.2024.26956,
    author = {Holmgren, A. Jay and Hendrix, Nathaniel and Maisel, Natalya and Everson, Jordan and Bazemore, Andrew and Rotenstein, Lisa and Phillips, Robert L. and Adler-Milstein, Julia},
    title = {Electronic Health Record Usability, Satisfaction, and Burnout for Family Physicians},
    journal = {JAMA Network Open},
    volume = {7},
    number = {8},
    pages = {e2426956-e2426956},
    year = {2024},
}

@article{Sasseville2025Impact,
    author = {Sasseville, Maxime and Yousefi, Farzaneh and Ouellet, Steven and Naye, Florian and Stefan, Theo and Carnovale, Val{\' e}rie and Bergeron, Fr{\' e}d{\' e}ric and Ling, Linda and Gheorghiu, Bobby and Hagens, Simon and Gareau-Lajoie, Samuel and Leblanc, Annie},journal = {Healthcare},doi = {10.3390/healthcare13121447},number = {12},year = {2025},month = {jun 16},pages = {1447--1447},title = {The {Impact} of {AI} {Scribes} on {Streamlining} {Clinical} {Documentation}: A {Systematic} {Review}},volume = {13},
}

@article{Miksanek2020Productivity,
author = {Miksanek, Tyler J and Skandari, M. Reza and Ham, Sandra A and Lee, Wei-Wei and Press, Valerie G. and Brown, Marie T and LAITEERAPONG, NEDA},journal = {Annals of Internal Medicine},doi = {10.7326/m20-0428},number = {1},year = {2020},month = {oct 5},pages = {1--7},title = {The {Productivity} {Requirements} of {Implementing} a {Medical} {Scribe} {Program}},volume = {174},
}

@article{Mess2025Artificial,author = {Mess, Sarah A. and Mackey, Alison J. and Yarowsky, David E.},journal = {Plastic \& Reconstructive Surgery Global Open},doi = {10.1097/gox.0000000000006450},number = {1},year = {2025},month = {jan 1},pages = {e6450--e6450},title = {Artificial {Intelligence} {Scribe} and {Large} {Language} {Model} {Technology} in {Healthcare} {Documentation}: Advantages, {Limitations}, and {Recommendations}},volume = {13},}

@article{Hassan2025Clinical,author = {Hassan, Hadeel and Zipursky, Amy R and Rabbani, Naveed and You, Jacqueline G. and Tse, Gabe and Orenstein, Evan and Ray, Mondira and Parsons, Chase and Shin, Stella and Lawton, Gregory and Jessa, Karim and Sung, Lillian and Yan, Adam P},journal = {Applied Clinical Informatics},doi = {10.1055/a-2597-2017},number = {04},year = {2025},month = {apr 30},pages = {1121--1135},title = {Clinical {Implementation} of {Artificial} {Intelligence} {Scribes} in {Health} {Care}: A {Systematic} {Review}},volume = {16},}

@article{H2021Current,
author = {Hämäläinen, Mika and Alnajjar, Khalid},journal = {arXiv (Cornell University)},year = {2021},month = {sep 23},title = {The {Current} {State} of {Finnish} {NLP}},
}

@article{Adedeji2024Sound,author = {Adedeji, Ayo and Joshi, Sarita and Doohan, Brendan},journal = {arXiv (Cornell University)},doi = {10.48550/arxiv.2402.07658},year = {2024},month = {feb 13},title = {The {Sound} of {Healthcare}: Improving {Medical} {Transcription} {ASR} {Accuracy} with {Large} {Language} {Models}},}

@article{Garcia2024Medical,author = {Garcia-Ferrero, Iker and Agerri, Rodrigo and Salazar, Aitziber Atutxa and Cabrio, Elena and de la Iglesia, Iker and Lavelli, Alberto and Magnini, Bernardo and Molinet, Benjamin and Ramirez-Romero, Johana and Rigau, German and Villa-Gonzalez, Jose Maria and Villata, Serena and Zaninello, Andrea},journal = {arXiv (Cornell University)},doi = {10.48550/arxiv.2404.07613},year = {2024},month = {apr 12},title = {Medical {mT5}: An {Open}-{Source} {Multilingual} {Text}-to-{Text} {LLM} for {The} {Medical} {Domain}},}

@article{Khatim2024Using,author = {Khatim, Nur Ahmad and Irfan, Azmul Asmar and Arief, Mansur M.},journal = {arXiv (Cornell University)},doi = {10.48550/arxiv.2409.17054},year = {2024},month = {sep 26},title = {Using {LLM} for {Real}-{Time} {Transcription} and {Summarization} of {Doctor}-{Patient} {Interactions} into {ePuskesmas} in {Indonesia}: A {Proof}-of-{Concept} {Study}},}

@article{Yang2025Fine,author = {Yang, Qimin and Chen, Jiexin and Sun, Yue and Wang, Yapeng and Tan, Tao},journal = {Quantitative Imaging in Medicine and Surgery},doi = {10.21037/qims-2024-2655},number = {6},year = {2025},month = {jun 1},pages = {5450--5462},title = {Fine-tuning medical language models for enhanced long-contextual understanding and domain expertise},volume = {15},}

@article{Vrbancic2020Transfer,author = {Vrban{\v c}i{\v c}, Grega and Podgorelec, Vili},journal = {IEEE Access},doi = {10.1109/access.2020.3034343},year = {2020},month = {jan 1},pages = {196197--196211},title = {Transfer {Learning} {With} {Adaptive} {Fine}-{Tuning}},volume = {8},}

@article{Wu2025LLM,author = {Wu, Xiao-Kun and Chen, Min and Li, Wanyi and Wang, Rui and Lu, Limeng and Liu, Jia and Hwang, Kai and Hao, Yixue and Pan, Yan-Ru and Meng, Qingguo and Huang, Kaibin and Hu, Long and Guizani, Mohsen and Chao, Naipeng and Fortino, Giancarlo and Lin, Fei and Tian, Yonglin and Niyato, Dusit and Wang, Fei-Yue},journal = {Big Data and Cognitive Computing},doi = {10.3390/bdcc9040087},number = {4},year = {2025},month = {apr 2},pages = {87--87},title = {LLM {Fine}-{Tuning}: Concepts, {Opportunities}, and {Challenges}},volume = {9},}

@article{Pol2024Hugging,author = {Pol, Urmila R.},journal = {International Journal for Research in Applied Science and Engineering Technology},doi = {10.22214/ijraset.2024.64023},number = {8},year = {2024},month = {aug 27},pages = {1121--1124},title = {Hugging {Face}: Revolutionizing {AI} and {NLP}},volume = {12},}

@inproceedings{Ait2023HFCommunity,
  author    = {Ait, Adem and Izquierdo, Javier Luis C{\' a}novas and Cabot, Jordi},
  title     = {{HFCommunity}: A {Tool} to {Analyze} the {Hugging} {Face} {Hub} {Community}},
  booktitle = {2023 IEEE International Conference on Software Analysis, Evolution and Reengineering (SANER)},
  year      = {2023},
  pages     = {728--732},
  month     = {mar},
  address   = {Taipa, Macao},
  publisher = {IEEE},
  doi       = {10.1109/saner56733.2023.00080}
}

@article{Susnjak2025Automating,author = {Susnjak, Teo and Hwang, Peter and Reyes, Napoleon and Barczak, Andre L. C. and McIntosh, Timothy and Ranathunga, Surangika},journal = {ACM Transactions on Knowledge Discovery from Data},doi = {10.1145/3715964},number = {3},year = {2025},month = {jan 31},pages = {1--39},title = {Automating {Research} {Synthesis} with {Domain}-{Specific} {Large} {Language} {Model} {Fine}-{Tuning}},volume = {19},}

@article{Shen2023HuggingGPT,author = {Shen, Yongliang and Song, Kaitao and Xu, Tan and Dongsheng, Li and Weiming, Lu and Zhuang, Yueting},journal = {arXiv (Cornell University)},doi = {10.48550/arxiv.2303.17580},year = {2023},month = {mar 31},title = {HuggingGPT: Solving {AI} {Tasks} with {ChatGPT} and its {Friends} in {Hugging} {Face}},}

@article{Zhiqiang2023LLM,author = {Zhiqiang, Hu and Lei, Wang and Lan, Yihuai and Xu, Wanyu and Lim, Ee-Peng and Bing, Lidong and Xing, Xu and Poria, Soujanya and Lee, Roy Ka-Wei},journal = {arXiv (Cornell University)},doi = {10.48550/arxiv.2304.01933},year = {2023},month = {apr 5},title = {LLM-{Adapters}: An {Adapter} {Family} for {Parameter}-{Efficient} {Fine}-{Tuning} of {Large} {Language} {Models}},}

@article{Lankford2023adaptMLLM,author = {Lankford, S{\' e}amus and Afli, Haithem and Way, Andy},journal = {Information},doi = {10.3390/info14120638},number = {12},year = {2023},month = {nov 29},pages = {638--638},title = {adaptMLLM: Fine-{Tuning} {Multilingual} {Language} {Models} on {Low}-{Resource} {Languages} with {Integrated} {LLM} {Playgrounds}},volume = {14},}

@article{Chen2024Meta,author = {Chen, Yaqi and Zhang, Hao and Yang, Xukui and Zhang, Wen-lin and Qu, Dan},journal = {Neurocomputing},doi = {10.1016/j.neucom.2024.128493},year = {2024},month = {aug 28},pages = {128493--128493},title = {Meta-{Adaptable}-{Adapter}: Efficient adaptation of self-supervised models for low-resource speech recognition},volume = {609},}

@article{Maity2025Large,author = {Maity, Subhankar and Saikia, Manob Jyoti},journal = {Bioengineering},doi = {10.3390/bioengineering12060631},number = {6},year = {2025},month = {jun 10},pages = {631--631},title = {Large {Language} {Models} in {Healthcare} and {Medical} {Applications}: A {Review}},volume = {12},}

@article{Hugo2023LLaMA,author = {Hugo, Touvron and Lavril, Thibaut and Izacard, Gautier and Martinet, Xavier and Lachaux, Marie-Anne and Lacroix, Timoth{\' e}e and Rozi{\` e}re, Baptiste and Goyal, Naman and Hambro, Eric and Azhar, Faisal and Rodriguez, Aurelien and Joulin, Armand and Grave, Edouard and Lample, Guillaume},journal = {arXiv (Cornell University)},doi = {10.48550/arxiv.2302.13971},year = {2023},month = {feb 28},title = {LLaMA: Open and {Efficient} {Foundation} {Language} {Models}},}

@misc{amd2023finnishai,
  author       = {{AMD}},
  title        = {SiloGen and Tietoevry Care are developing a Finnish-speaking AI assistant for healthcare professionals},
  year         = {2023},
  howpublished = {\href{https://www.amd.com/en/blogs/2023/silogen-and-tietoevry-care-are-developing-a-finnish-speaking-ai-assistant-for-healthcare-professionals.html}{AMD blog post}},
  note         = {Accessed: 2026-03-22}
}

@article{Mehdi2024Teuken,author = {Mehdi, Ali and Fromm, Michael and Thellmann, Klaudia and Ebert, Jan and Weber, Alexander Arno and Rutmann, Richard and Jain, Charvi and L{\" u}bbering, Max and Steinigen, Daniel and Leveling, Johannes and Klug, Katrin and Buschhoff, Jasper Schulze and Jurkschat, Lena and Hammam, Abdelwahab and Stein, Benny J{\" o}rg and Sylla, Karl-Heinz and Denisov, Pavel and Brandizzi, Nicolo' and Saleem, Qasid and Bhowmick, Anirban and Helmer, Lennard and John, Chelsea and Suarez, Pedro Ortiz and Ostendorff, Malte and Jude, Alex and Manjunath, Lalith and Weinbach, Samuel and Penke, Carolin and Filatov, Oleg and Barth, Fabio and Paramita, Mirza and Weber, Lucas and Ines, Wendler and Sifa, Rafet and Fabian, Kuch and Herten, Andreas and J{\" a}kel, Ren{\' e} and Rehm, Georg and Kesselheim, Stefan and K{\" o}hler, Joachim and Flores-Herr, Nicolas},journal = {arXiv (Cornell University)},doi = {10.48550/arxiv.2410.03730},year = {2024},month = {oct 8},title = {Teuken-7B-{Base} \& {Teuken}-7B-{Instruct}: Towards {European} {LLMs}},}

@article{Acikgoz2024Hippocrates,author = {Acikgoz, Emre Can and {\. I}nce, Osman Batur and Bench, Rayene and Boz, Arda An\i{}l and Kesen, {\. I}lker and Erdem, Aykut and Erdem, Erkut},journal = {arXiv (Cornell University)},doi = {10.48550/arxiv.2404.16621},year = {2024},month = {apr 26},title = {Hippocrates: An {Open}-{Source} {Framework} for {Advancing} {Large} {Language} {Models} in {Healthcare}},}

@article{2024DeepSeek,author = {Xiao, Bi and Deli, Chen and Guanting, Chen and Chen, Shanhuang and Dai, Damai and Cheng-qi, Deng and Hong-hui, Ding and Kai, Dong and {Du Qiushi} and Zhe, Fu and Gao, Huazuo and Kaige, Gao and Wenjun, Gao and Ge, Ruiqi and Kang, Guan and Guo, Daya and Jian-zhong, Guo and Guang-bo, Hao and Zhe-wen, Hao and Ying, He and Hu, Wenjie and PanPan, Huang and Li, Erhang and Guowei, Li and Jia-Shi, Li and Yao, Li and , Li Y K and Wenfeng, Liang and Lin, Fangyun and Liu, A. X. and Bo, Liu and Wen, Liu and Xiao-dong, Liu and Xin, Liu and Yiyuan, Liu and Lu, Haoyu and Lu, Shanghao and Luo, Fuli and ShiRong, Ma and Xiaotao, Nie and Tian, Pei and Piao, Yishi and Junjie, Qiu and Hui, QU and Ren, Tongzheng and Ren, Zehui and Chong, Ruan and Sha, Zhangli and Zhi-hong, Shao and Junxiao, Song and Xuecheng, Su and Sun, Jingxiang and Yaofeng, Sun and Minghui, Tang and Bingxuan, Wang and Peiyi, Wang and Wang, Shiyu and Yao-hui, Wang and Yong-Ji, Wang and Wu, Tong and , Wu Y and Xin, Xie and Xie, Zhenda and Xie, Ziwei and Yi-liang, Xiong and Hanwei, Xu and Xu, R. X. and Yanhong, Xu and Dejian, Yang and Yu-xiang, You and Shuiping, Yu and Xingkai, Yu and , Zhang B and Haowei, Zhang and Zhang, Lecong and Li-yue, Zhang and Ming-Chuan, Zhang and Ming-hua, Zhang and Zhang, Wentao and Yi-chao, Zhang and Chenggang, Zhao and Zhao, Yao and Shang-yan, Zhou and Shunfeng, Zhou and Zhu, Qihao and Yuheng, Zou},journal = {arXiv (Cornell University)},doi = {10.48550/arxiv.2401.02954},year = {2024},month = {jan 8},title = {DeepSeek {LLM}: Scaling {Open}-{Source} {Language} {Models} with {Longtermism}}
}

@article{virtanen2019multilingual,
  title     = {Multilingual is not enough: BERT for Finnish},
  author    = {Virtanen, Antti and Kanerva, Jenna and Ilo, Rami and Luoma, Jouni and Luotolahti, Juhani and Salakoski, Tapio and Ginter, Filip and Pyysalo, Sampo},
  journal   = {arXiv preprint arXiv:1912.07076},
  year      = {2019},
  url       = {https://arxiv.org/abs/1912.07076}
}

@article{Luo2022BioGPT,author = {Luo, Renqian and Sun, Liai and Xia, Yingce and Qin, Tao and Sheng, Zhang and Poon, Hoifung and Liu, Tie-Yan},journal = {Briefings in Bioinformatics},doi = {10.1093/bib/bbac409},number = {6},year = {2022},month = {sep 24},title = {BioGPT: generative pre-trained transformer for biomedical text generation and mining},volume = {23},}

@article{Matondora2024NLP,author = {Matondora, L and Mutandavari, M and Mupini, B},journal = {International Journal of Innovative Science and Research Technology (IJISRT)},doi = {10.38124/ijisrt/ijisrt24jul1191},year = {2024},month = {aug 10},pages = {2549--2557},title = {NLP {Based} {Prediction} of {Hospital} {Readmission} using {ClinicalBERT} and {Clinician} {Notes}},}

@article{Tianyi2022BERTScore,author = {Tianyi, Zhang and Kishore, Varsha and Wu, Felix and Weinberger, Kilian Q. and Artzi, Yoav},journal = {arXiv (Cornell University)},doi = {10.48550/arxiv.1904.09675},year = {2022},month = {feb 28},title = {BERTScore: Evaluating {Text} {Generation} with {BERT}},}

@article{Glushkova2023BLEU,author = {Glushkova, Taisiya and Zerva, Chrysoula and Martins, Andr{\' e} F. T.},journal = {arXiv (Cornell University)},doi = {10.48550/arxiv.2305.19144},year = {2023},month = {may 31},title = {BLEU {Meets} {COMET}: Combining {Lexical} and {Neural} {Metrics} {Towards} {Robust} {Machine} {Translation} {Evaluation}},}

@article{Barbella2022Rouge,author = {Barbella, Marcello and Tortora, Genoveffa},journal = {SSRN Electronic Journal},doi = {10.2139/ssrn.4120317},year = {2022},month = {jan 1},title = {Rouge {Metric} {Evaluation} for {Text} {Summarization} {Techniques}},}

@article{Tang2024Improving,author = {Tang, Gongbo and Yousuf, Oreen and Jin, Zeying},journal = {IEEE Access},doi = {10.1109/access.2024.3406993},year = {2024},month = {jan 1},pages = {77739--77749},title = {Improving {BERTScore} for {Machine} {Translation} {Evaluation} {Through} {Contrastive} {Learning}},volume = {12},}

@article{Department2021Performance,author = {{Department of Computer Science and Informatics, University of Energy and Natural Resources} and Nti, Isaac Kofi and Boateng, Owusu Nyarko and Aning, Justice},journal = {International Journal of Information Technology and Computer Science},doi = {10.5815/ijitcs.2021.06.05},number = {6},year = {2021},month = {dec 8},pages = {61--71},title = {Performance of {Machine} {Learning} {Algorithms} with {Different} {K} {Values} in {K}-fold {CrossValidation}},volume = {13},}

@article{Lin2022Curriculum,author = {Lin, Zeyang and Lai, Jun and Chen, Xiliang and Cao, Lei and Wang, Jun},journal = {Entropy},doi = {10.3390/e24121787},number = {12},year = {2022},month = {dec 6},pages = {1787--1787},title = {Curriculum {Reinforcement} {Learning} {Based} on {K}-{Fold} {Cross} {Validation}},volume = {24},}

@misc{csc_puhti,
  title        = {Puhti Supercomputer},
  author       = {{CSC -- IT Center for Science}},
  year         = {2025},
  howpublished = {\url{https://docs.csc.fi/computing/systems-puhti/}},
  note         = {Accessed: 2025-05-18},
  address      = {Espoo, Finland},
  publisher    = {CSC -- IT Center for Science}
}

@misc{python3115,
  title        = {Python Version 3.11.5},
  author       = {{Python Software Foundation}},
  year         = {2023},
  howpublished = {\url{https://www.python.org/}},
  note         = {Accessed: 2026-03-22}
}

@misc{cuda117,
  title        = {CUDA Toolkit 11.7},
  author       = {{NVIDIA Corporation}},
  year         = {2022},
  howpublished = {\url{https://developer.nvidia.com/cuda-toolkit}},
  note         = {Accessed: 2026-03-22}
}

@misc{metropolia_innovation_2025,
  author       = {{Metropolia University of Applied Sciences}},
  title        = {Innovation Projects},
  year         = {2025},
  howpublished = {\url{https://www.metropolia.fi/en/rdi/innovation-projects}},
  address      = {Helsinki, Finland},
  note         = {Accessed: 2025-05-11}
}

\end{document}